\documentclass{article}
\usepackage{mathpazo}
\usepackage[margin=1in]{geometry}
\usepackage[style=alphabetic,natbib=true]{biblatex}
\usepackage{microtype}
\usepackage{graphicx}
\usepackage{caption}
\usepackage{subcaption}
\usepackage{booktabs} %

\usepackage{hyperref}

\usepackage{algorithm}
\usepackage{algorithmic}

\usepackage{amsmath}
\usepackage{amssymb}
\usepackage{mathtools}
\usepackage{amsthm}
\usepackage{bbold}

\usepackage[capitalize,noabbrev]{cleveref}
\usepackage{multirow}
\usepackage{enumitem}
\usepackage{tikz}
\usepackage{pgfplots}

\theoremstyle{plain}

\theoremstyle{definition}

\theoremstyle{remark}

\usepackage[textsize=tiny]{todonotes}

\renewrobustcmd{\bfseries}{\fontseries{b}\selectfont}
\renewrobustcmd{\boldmath}{}
\newrobustcmd{\B}{\bfseries}
\newcommand{\dice}{DiCE}
\newcommand{\ceds}{shift explanation}
\newcommand{\ourmethod}{GSE}

\newcommand{\psrc}{\ensuremath{P}}
\newcommand{\ptgt}{\ensuremath{Q}}

\newcommand{\civil}{Civil Comments}

\DeclareMathOperator*{\argmin}{arg\,min}
 \title{Rectifying Group Irregularities in Explanations for Distribution Shift}
\author{
    Adam Stein \qquad Yinjun Wu \qquad Eric Wong \qquad Mayur Naik\\
    {\normalsize \textsc{\{steinad, wuyinjun, exwong, mhnaik\}@seas.upenn.edu}} \\
    University of Pennsylvania
}
\date{}

\begin{document}
\maketitle

\begin{abstract}
It is well-known that real-world changes constituting distribution shift adversely affect model performance.
How to characterize those changes in an interpretable manner is poorly understood.
Existing techniques to address this problem take the form of shift explanations that elucidate how to map samples from the original distribution toward the shifted one by reducing the disparity between these two distributions.
However, these methods can introduce group irregularities, leading to explanations that are less feasible and robust. 
To address these issues, we propose Group-aware Shift Explanations (\ourmethod{}), a method that produces interpretable explanations by leveraging worst-group optimization to rectify group irregularities. We demonstrate how \ourmethod{} not only maintains group structures, such as demographic and hierarchical subpopulations, but also enhances feasibility and robustness in the resulting explanations in a wide range of tabular, language, and image settings.
\end{abstract}

\section{Introduction}\label{sec: intro}
Classic machine learning theory assumes that the training and testing data are sampled from the same distribution.
Unfortunately, distribution shifts infringe on this requirement and can drastically change a model's behavior \citep{kurakin2018adversarial}.
For instance, training a model on data collected from one hospital may result in inaccurate diagnoses for patients from other hospitals due to variations in medical equipment \citep{zech2018variable, subbaswamy2020development}. Similarly, shifts from daytime to nighttime or from clear to rainy weather are major obstacles for autonomous driving \citep{dai2018dark, wang2020pedestrian}.

When such a distribution shift occurs, it is often useful to understand \emph{why} and \emph{how} the data changed. For example, suppose a doctor observes that their medical AI model's performance is degrading. Before arbitrarily changing the model, the doctor should first understand the changes in their patient data \citep{subbaswamy2020development}. Similarly, a self-driving engineer would have an easier time to adapt an autonomous driving system to a new environment if it was known that the shift resulted from changing weather conditions \citep{manot}. In addition, policymakers need to understand why and how an event or crisis happens so that they can adjust their policies appropriately \citep{lundgren2018stability}. 

To facilitate better understanding of a distribution shift, it is crucial to generate appropriate {\em shift explanations}. The format of a shift explanation is a mapping from the original distribution (called source distribution) to the shifted one (called target distribution) such that their disparity is reduced. For example, \citet{kulinski2022towards} find a direct mapping of points from the original distribution toward the shifted one via optimal transport \citep{peyre2017computational} and its variant, $K$-cluster transport. Another approach is to use counterfactual explanation methods such as \dice{} \citep{mothilal2020explaining} which explain classifiers. A counterfactual explanation of a classifier between source and target distributions will map each source instance such that the model classifies this instance as from the target distribution.

Shift explanations produced by state-of-the-art methods seek to optimize global objectives such as minimizing the difference between the target distribution and the mapped source distribution.
However, mappings that merely satisfy this goal are not necessarily good explanations: they can fail to be feasible in practice, or lack robustness to perturbations in the source distribution.
This in turn fundamentally limits the practicality of shift explanations produced by existing methods.

\begin{figure}
    \centering
    \begin{subfigure}[b]{0.6\textwidth}
        \includegraphics[width=\linewidth]{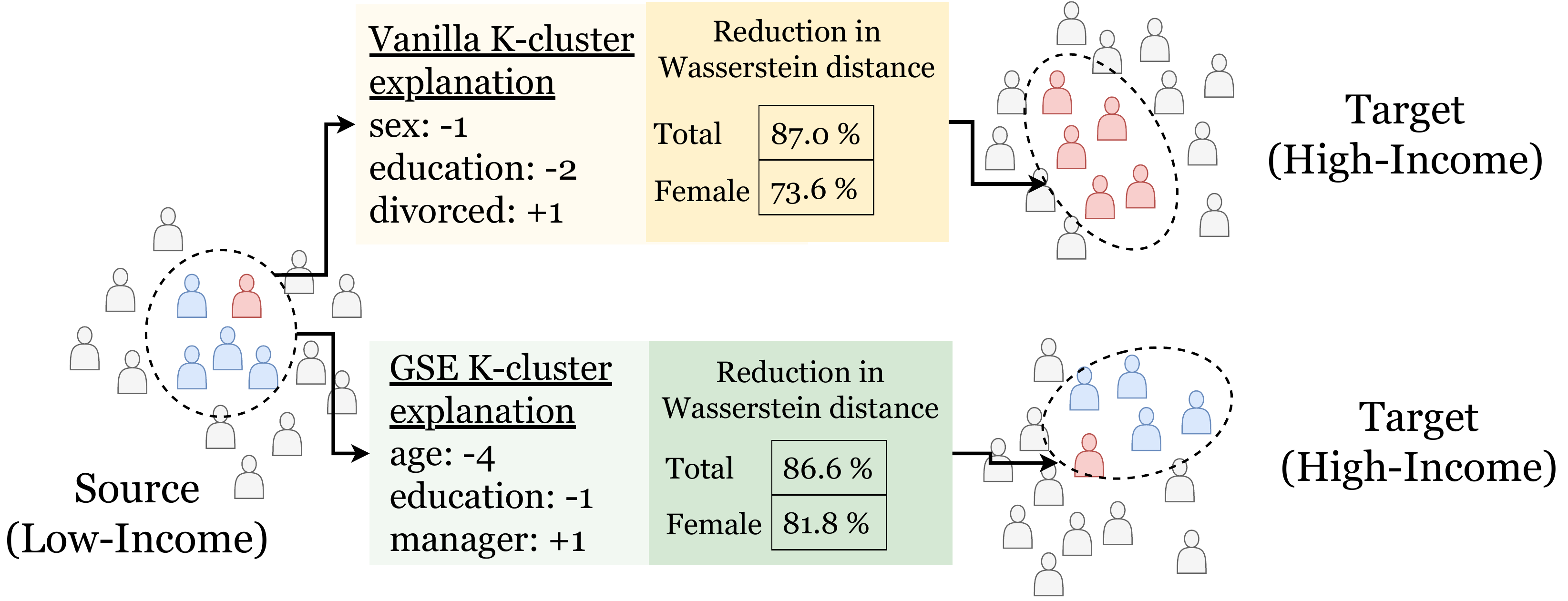}
        \caption{}
        \label{fig:overview-1}
    \end{subfigure}%
    \hfill
    \begin{subfigure}[b]{0.36\textwidth}
        \includegraphics[width=\linewidth]{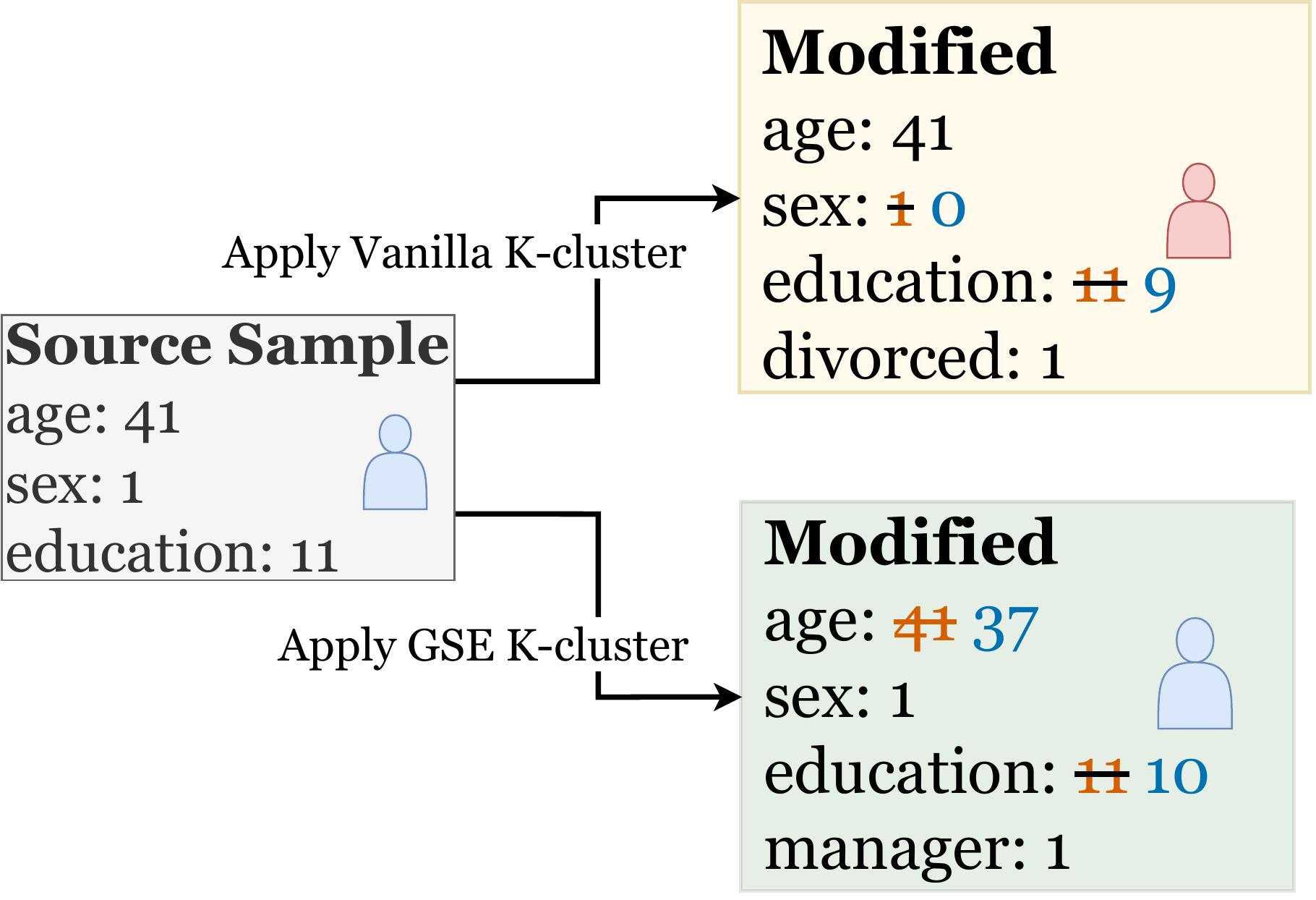}
        \caption{}
        \label{fig:overview-2}
    \end{subfigure}%
    \vspace{-0.05in}
    \caption{(a) shows an example of explaining the distribution shift from a source (i.e., low-income) population to a target (i.e., high-income) population in a subset of the Adult dataset \cite{blake1998uci} using two different methods: vanilla $K$-cluster transport \citep{kulinski2022towards} and our \ourmethod{} $K$-cluster transport. 
    The shift explanation produced by the vanilla method explains the shift by subtracting 1 from the ``sex'' attribute.
    On the other hand, \ourmethod{} generates an explanation that preserves the male and female subpopulations, and instead requires the samples in the source to change their education level and age.
    (b) shows how a particular member of the cluster is modified under the two methods. The explanation produced by the state-of-the-art method changes the sample's sex whereas our method does not.
    }
    \label{fig:overview}
    \vspace{-0.25in}
\end{figure}

As a concrete example of this phenomenon, Figure \ref{fig:overview} shows explanations from \citep{kulinski2022towards} that map individuals with low income (source distribution) to individuals with high income (target distribution) in the Adult dataset from the UCI Machine Learning Repository \citep{blake1998uci}.
Such explanations can help reveal insights about income inequalities that enable a policymaker to propose better policies or an individual to understand how to increase their income.
At a dataset level,  $K$-cluster transport \citep{peyre2017computational} can produce a shift explanation that effectively maps the source distribution to the target, resulting in a 87\% reduction in the Wasserstein distance between these two distributions.
However, upon closer inspection, this explanation shifts a majority male cluster to a majority female cluster.
Focusing on the female subpopulation of the source and target, the explanation only decreases the Wasserstein distance by 73.6\%.
Such an explanation is not useful if gender change is infeasible---or less feasible than changing other attributes such as education level.

Our key insight to achieving high-quality shift explanations is to steer the generated explanations to respect subpopulations, or {\bf groups}, in the data. 
Since groups are highly context-specific in nature, we seek an approach that is general and factors this objective jointly with the overall fitness of the produced mapping
from source to target populations.
In our running example, assuming gender-based grouping, such an approach should yield a mapping that minimizes disrupting the groups while maximizing overall fitness.
As depicted at the bottom of Figure \ref{fig:overview-1}, it is in fact possible to achieve such a mapping by using the \textit{same} underlying K-cluster transport method, that increases the reduction of Wasserstein distance from 73.6\% to 81.8\% between source samples and target samples within the female group, and only slightly impacts the reduction of Wasserstein distance between the overall source and target populations (from 87\% to 86.6\%).

To this end, we propose Group-aware Shift Explanations (\ourmethod{}), an explanation method for distribution shift that preserves groups in the data.
We develop a unifying framework that incorporates heterogeneous methods for producing shift explanations in diverse settings and allows us to apply \ourmethod{} to these methods.
In addition, \ourmethod{} enhances the feasibility and robustness of the resulting explanations. Through extensive experiments over a wide range of tabular, language, and image datasets, we demonstrate that \ourmethod{} not only maps source samples closer to target samples belonging to the same group, thus preserving group structure, but also boosts the feasibility and robustness by up to 28\% and 
42\% respectively. 

Our main contributions are summarized as follows:

\begin{enumerate}[leftmargin=*,itemsep=0pt,topsep=0pt,parsep=0pt]
    \item We identify and demonstrate group irregularities as a class of problems that can adversely affect the quality of shift explanations by state-of-the-art methods.
    \item We propose one method, Group-aware Shift Explanations (GSE), to rectify group irregularities when explaining distribution shift.%
    \item We propose a general framework to unify heterogeneous shift explanation methods such as $K$-cluster transport and show how they can be integrated into \ourmethod{} to enforce group structures across varied settings, including tabular, NLP, and image settings. %
    \item We demonstrate how \ourmethod{} maintains group structures and enables more feasible and robust shift explanations occurring in diverse datasets across different domains.%
\end{enumerate}

\vspace{-0.05in}
\section{Motivation}
\label{sec: problem}

In this section, we identify issues with existing shift explanations in terms of group irregularities.

\vspace{-0.05in}
\subsection{Constructing Mappings for Shift Explanations}
A distribution shift is any change from an initial distribution, called the source, to another distribution, called the target.
We follow prior work from \citet{kulinski2022towards} to define a shift explanation as a mapping from the source distribution to the target distribution. 
For instance, Figure~\ref{fig:overview} shows a shift explanation, called a $K$-cluster explanation \cite{kulinski2022towards} which maps the source distribution to the target distribution by subtracting 1 from the ``sex'' attribute among other changes.
Different shift explanation methods can produce different types of mappings. 

\vspace{-0.05in}
\subsection{Group Irregularities in Existing Shift Explanations}\label{sec: group_issue}
To find a shift explanation, state-of-the-art methods primarily minimize the disparity between the source distribution and the target distribution. For example, $K$-cluster transport minimizes an objective depending on the Wasserstein distance between the source and the target distribution \citep{peyre2017computational}. 
However, this is not sufficient for finding high-quality explanations.
Figure~\ref{fig:overview} shows such an example with $K$-cluster explanations where a mostly male group of the source gets mapped to a female group. 
In this case, the overall Wasserstein distance is reduced by 87\%, but the Wasserstein distance for the female subpopulation is decreased much less in Figure~\ref{fig:overview-1}. 

\vspace{-0.05in}
\subsubsection{Impact on Explanation Feasibility}
Shift explanations which break apart groups of the data are not only problematic because they degrade on the subpopulation level, but they can also be overall \textit{infeasible}. Feasibility is a measure of how useful an explanation is to a downstream user.
For instance, in Figure~\ref{fig:overview}, the sex attribute may be unactionable, so the $K$-cluster explanation which modifies the sex attribute would be useless for a policymaker 
who designs policies to help increase the income of the low-income population. Overall, the $K$-cluster explanation in Figure~\ref{fig:overview} is 
only feasible for 75.5\% of the source distribution, meaning that 24.5\% of the source samples have their sex attribute modified by the shift explanation. Later, we show how our method, which rectifies these group irregularities, results in more feasible explanations for the overall source distributions. 

\vspace{-0.05in}
\subsubsection{Impact on Explanation Robustness}

\begin{figure}[t]
    \centering
    \begin{subfigure}{0.39\textwidth}
    \includegraphics[width=\textwidth]{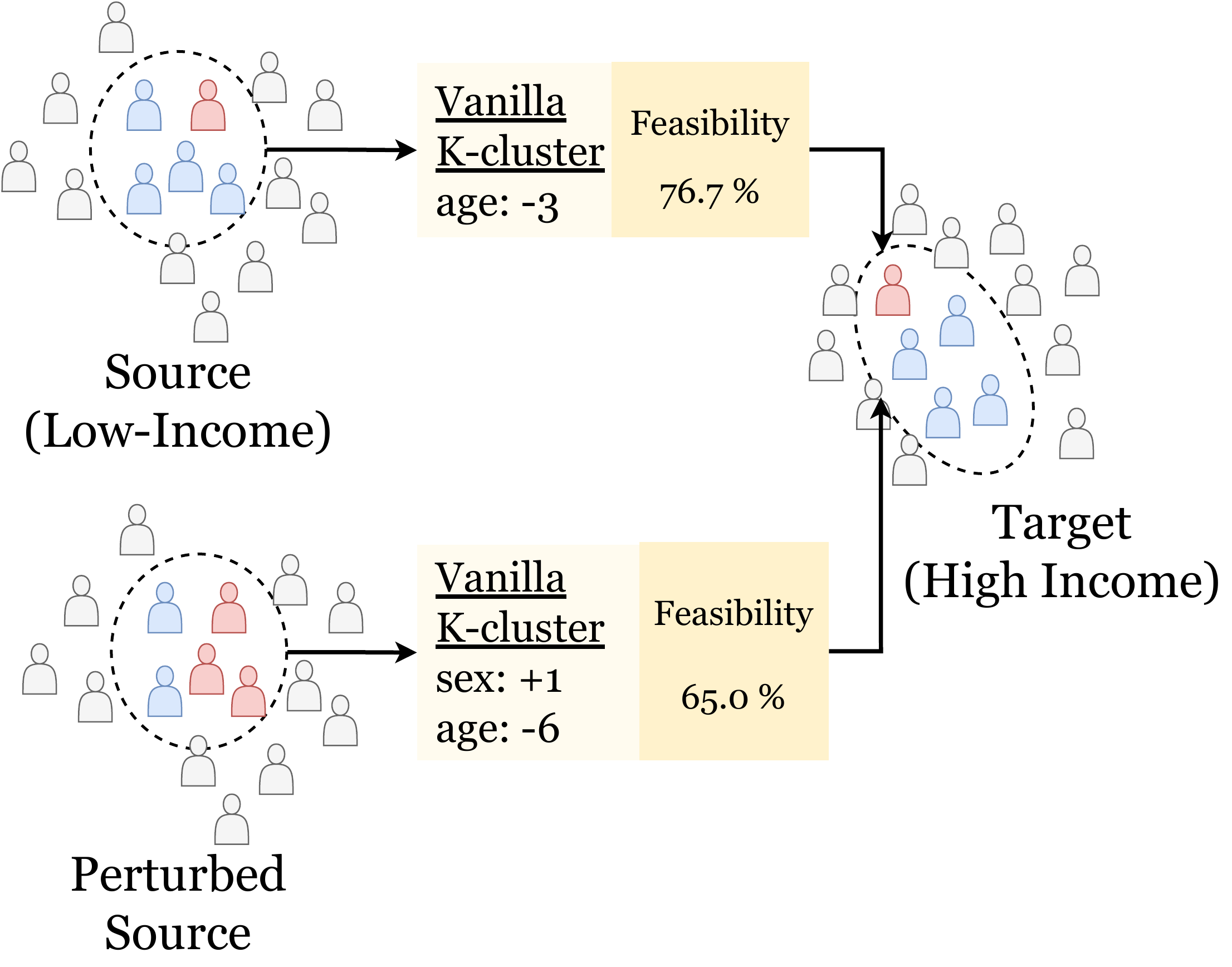}
    \caption{Adult dataset.}
    \label{fig:adult-robustness}
    \end{subfigure}%
    \hfill
    \begin{subfigure}{0.59\textwidth}
    \includegraphics[width=\textwidth]{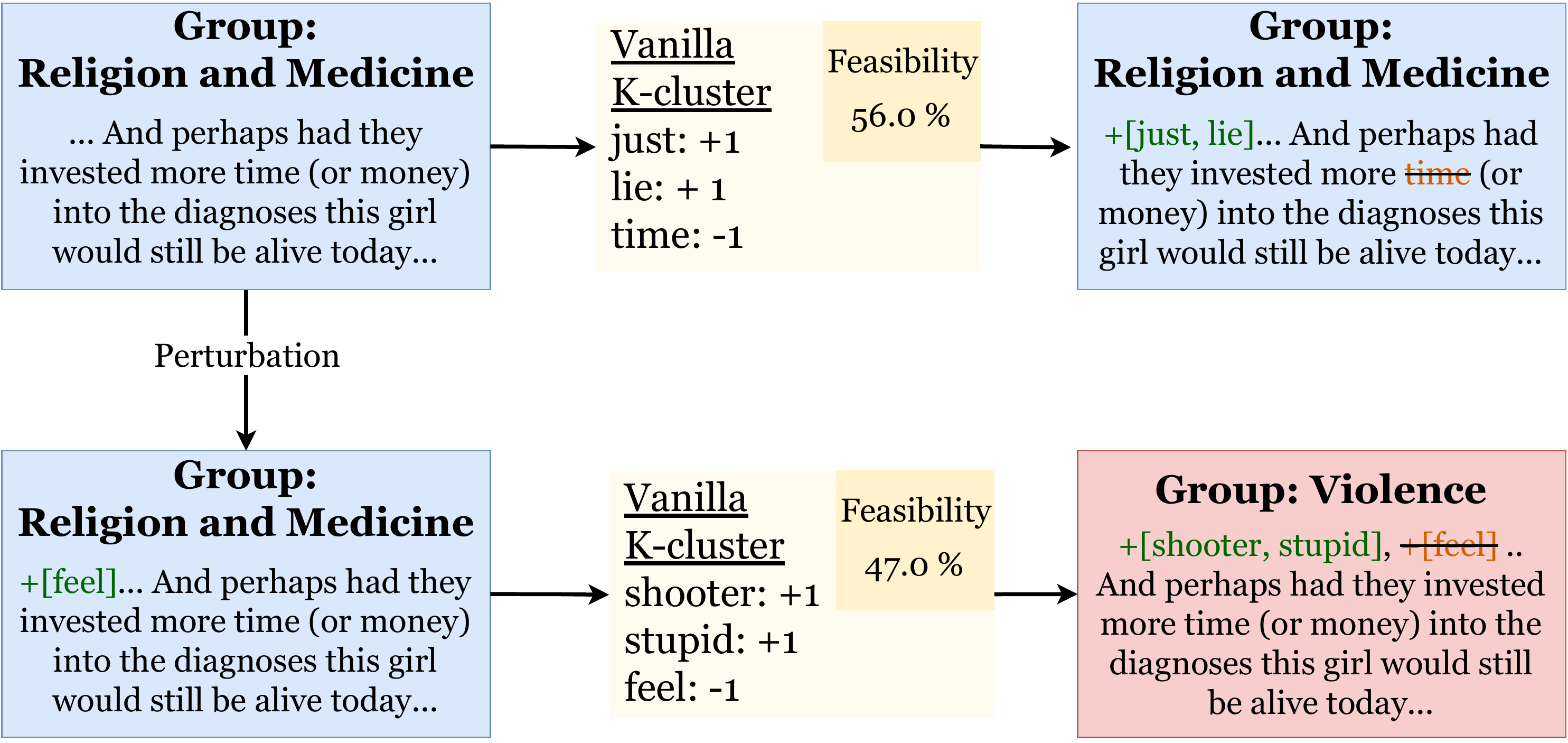}
    \caption{\civil{} dataset.}
    \label{fig:nlp-robustness}
    \end{subfigure}
    \caption{An example of poor robustness of an explanation. Even if an explanation is feasible (top), small perturbations to the source distribution can make it become infeasible (bottom). The details on how groups are derived for the \civil{} data is in Appendix~\ref{no-group-exp}.} %
    \label{fig:robustness-example}
\end{figure}
Group irregularities can also reduce \textit{robustness}, meaning that small changes to a source distribution result in large changes to the shift explanation.
Figure~\ref{fig:robustness-example} shows an example of poor explanation robustness in the Adult and \civil{} dataset. In Figure~\ref{fig:adult-robustness}, a small perturbation to the source distribution leads to the explanation changing from subtracting three from age to adding one to sex (changing from female to male) and subtracting six from age. Figure~\ref{fig:nlp-robustness} shows a shift explanation that maps a non-toxic sample relating to medicine into the target distribution of toxic sentences. After a small perturbation, the explanation maps the same sample by adding the words ``shooter'' and ``stupid'' which is an unfeasible change since it changes the topic of the sample to violence. Ideally, we want a shift explanation to be robust to very small changes to the source distribution since it should explain general behavior instead of relying on minute details of a distribution.

Robustness is problematic even when overall feasibility is high.
For example, in Figure~\ref{fig:adult-robustness}, the $K$-cluster explanation is feasible for 76.7\% of the source distribution samples.
After applying a small perturbation, however, the $K$-cluster explanation now modifies the sex attribute for an additional 11.7\% of the source distribution samples, reducing feasibility to 65.0\%.
Thus, even an explanation with high overall feasibility is not ideal if small changes to the source distribution can lead to drastic changes to the explanation.

\section{Group-aware Shift Explanations (\ourmethod{})}\label{sec: methods}

In this section, we discuss our method, Group-aware Shift Explanations (\ourmethod{}). First, we introduce \ourmethod{} in the context of $K$-cluster transport, and PercentExplained (PE) a Wasserstein-distance based metric for evaluating the quality of shift explanations. Then, we present a unified shift explanation framework which allows \ourmethod{} to work with arbitrary shift explanations and also generalizes it from tabular data to NLP and image data. Finally, we formalize the notions of feasibility and robustness introduced in Section~\ref{sec: problem} as additional metrics to evaluate the quality of shift explanations.

\subsection{Preliminaries on $K$-cluster transport and PercentExplained (PE)}

The shift explanations produced by $K$-cluster transport can be denoted by a mapping function $M(x; \theta_x)$. The function $M(x; \theta_x)$ maps a source sample $x$ towards the target distribution by a distance of $\theta_x$, which is a learnable parameter. As the name $K$-cluster transport suggests, all the source samples are grouped into a set of clusters, $C$, with $K$-means clustering algorithm, and within one cluster, $c\in C$, all the samples share the same $\theta_c$. Therefore, the mapping function for $K$-cluster transport is formulated as follows:
\begin{small}
\begin{align*}
    M(x; \theta) = x + \sum\nolimits_{c\in C}\mathbb{1}_{x\in c} \theta_c, \text{in which, } \theta = \{\theta_c | c \in C\}.
\end{align*}
\end{small}
{\bf Optimizing $\theta$ for $K$-cluster transport.}
According to \citep{kulinski2022towards}, 
$\theta$ is solved by maximizing PercentExplained (PE). Suppose the source distribution and the target distribution are denoted by $P$ and $Q$ respectively, then PE is formulated as follows:
\begin{small}
\begin{align}\label{eq: pe}
    \text{PE}(\theta; M, \psrc{}, \ptgt{}) = 1-W_2^2(M(\psrc{}; \theta), \ptgt{})/W_2^2(\psrc{}, \ptgt{}),
\end{align}
\end{small}
where $W_2(\cdot)$ is the Wasserstein-2 distance and $M(P; \theta)$ is the mapping $M$ applied to every sample in the source, $\{M(x; \theta)\mid x\in\psrc\}$.
Intuitively speaking, PE quantifies how much the distance between $\psrc{}$ and $\ptgt{}$ is reduced after $M(\cdot; \theta)$ is applied to $\psrc$.
A high PE means that the explanation closely matches the overall source to the overall target distribution. Using differentiable implementations of the Wasserstein-2 distance, like the GeomLoss library \citep{feydy2019interpolating}, allows us to directly optimize PE using gradient descent. 

\subsection{Worst-group PE for \ourmethod{}}
\label{sec: group_aware_objs}
To rectify the issues identified in Section~\ref{sec: problem} in existing shift explanations, we can ideally optimize PE for all pre-specified groups such that all groups are preserved by the shift explanation. This ideal, however, is not applicable to finding dataset-level explanations. Instead, we propose Group-aware Shift Explanations (\ourmethod{}) to optimize the {\it worst-group PE} among all groups, which thus implicitly improves PE for {\it all groups} simultaneously.

Specifically, suppose the source and target are partitioned into $G$ disjoint groups, i.e., $\psrc{} = \{\psrc_1, \psrc_2,\dots, \psrc_G\}$ and $\ptgt{} = \{\ptgt_1, \ptgt_2,\dots, \ptgt_G\}$, in which, $\psrc_g$ and $\ptgt_g$ belong to the same group, e.g., the male sub-populations from $\psrc$ and $\ptgt$. We can now evaluate PE on a shared group from the source and target as follows: 
\begin{small}
\begin{align}\label{eq: pe_sub}
    \text{PE}_{g}(\theta; M, \psrc_{g}, \ptgt_{g}) = 1-W_2^2(M(\psrc_{g}; \theta), \ptgt_{g})/{W_2^2(\psrc_{g}, \ptgt_{g})}.
\end{align}    
\end{small}
The above formula measures how much the distance between $\psrc_{g}$ and $\ptgt_{g}$ is reduced by the given shift explanation, $M$. Then worst-group PE can then be calculated over all $G$ groups, 
i.e.,:
\begin{small}
\begin{align}\label{eq: worst_group_pe}
    \text{WG-PE}(\theta; M, P, Q) = \min\nolimits_g \text{PE}_{g}(\theta; M, \psrc_g, \ptgt_g). 
\end{align}    
\end{small}
This metric indicates how much the distance between any pair of $\psrc_g$ and $\ptgt_g$ is reduced, in the {\em worst case}. Instead of learning a shift explanation which maximizes PE over the entire distributions but may leave some groups with arbitrarily small PE, \ourmethod{} learns an explanation maximizing WG-PE. Optimizing $\theta$ to maximize the WG-PE can guarantee that for {\em every} pair of $\psrc_g$ and $\ptgt_g$, $\text{PE}_{g}(\theta; \psrc_g, \ptgt_g, M)$ is not approaching 0. 

Intuitively, \ourmethod{} regularizes the groups where PE becomes arbitrarily small even though the overall PE is large.
Note that the goal of \ourmethod{} is still to learn shift explanations at the dataset level rather than find explanations for each group separately. 
This means that the vanila $K$-cluster transport and \ourmethod{} $K$-cluster transport produce explanations of the same complexity. As we will show in Section \ref{sec: exp}, both feasibility and robustness issues can also be mitigated with \ourmethod{}.

\subsection{A Unified Framework for General Settings}\label{sec: framework}
In this section, we propose a generic framework which generalizes \ourmethod{} from $K$-cluster transport to broad types of shift explanation methods, and from tabular data to a wide range of settings, including NLP and image data. 
\subsubsection{Generalizing to other shift explanation methods}

\paragraph{Generalizing the Mapping $M(x; \theta)$} First of all, recall that the shift explanations produced by $K$-cluster transport could be represented by the mapping function $M(x; \theta)$, which can be any function taking the sample $x\in\psrc$ and the moving distance $\theta$ as input. For example, for optimal transport \citep{kulinski2022towards}, $M(x; \theta) = x + \theta_x$ where the moving distance, $\theta_x$, varies between different $x$. 
\paragraph{Generalizing the Objective Function beyond PE}
Note that one objective function, PercentExplain (PE), is optimized for solving $\theta$ for $K$-cluster transport. Indeed, any differentiable loss function, $L(\theta; M, \psrc{}, \ptgt{})$ for optimizing $\theta$ can be employed, 
which takes the mapping, $M$; the source distribution, $\psrc$; and the target distribution, $\ptgt$, as input.
For instance, for optimal transport and $K$-cluster transport \citep{kulinski2022towards}, 
$L$ is $1 - \text{PE}$.
The details for how to instantiate $M$ and $L$ for other shift explanation methods, e.g., optimal transport and \dice{}, are given in Appendix~\ref{appendix: instantiations}. But note that the feasibility and robustness metrics introduced in Section \ref{sec: group_issue} (will be formalized in Section \ref{sec: fr}) are not suitable due to their non-differentiability. Therefore, they only serve as post-hoc evaluation metrics.

We can now provide a general form of \ourmethod{} for any shift explanation method decomposed as a parameterized mapping $M(\cdot; \theta)$ and an objective function $L$ for learning $\theta$.
First, we extend our formulation of WG-PE in Equation~\ref{eq: worst_group_pe} beyond PE by replacing PE with $1-L$ (recall that $L$ is $1 - \text{PE}$ for $K$-cluster transport), i.e:
\begin{small}
\begin{align}\label{eq: group_loss}
\begin{split}
    & \text{WG-}L(\theta; M, \psrc{}, \ptgt{}) = \min\nolimits_g(\{1 - L(\theta; M, \psrc_g, \ptgt_g)\}_{g=1}^G) = \max\nolimits_g(\{L(\theta; M, \psrc_g, \ptgt_g)\}_{g=1}^G)
\end{split}
\end{align}    
\end{small}
Recall that $\psrc_g$ and $\ptgt_g$ represent a group of samples from the source and the target respectively, belonging to the same group. 
We further generalize Equation \eqref{eq: group_loss} by using an arbitrary aggregation function $F$ in place of the $\max$ function and regularizing with the loss calculated between the whole $\psrc{}$ and $\ptgt{}$ to balance the optimization between the worst group and the overall distribution, i.e.:
\begin{small}
\begin{align}\label{eq: generalized_loss}
\begin{split}
    & \text{WG-}L(\theta; M, \psrc{}, \ptgt{}) = \argmin\nolimits_\theta \left( F(\{L_g(\theta; M, \psrc_g, \ptgt_g)\}_{g=1}^G) + \lambda \cdot L(\theta; M, \psrc{}, \ptgt{}) \right)
\end{split}
\end{align}    
\end{small}
where $\lambda$ is a hyper-parameter and $F$ is an aggregation function. The choice of $F$ and $\lambda$ for our experiments is given in Appendix~\ref{app: framework-hyp}.

\subsubsection{Generalizing to language and image data}
\label{lang-img-setup}

\begin{figure*}[t]
    \centering
    \includegraphics[width=\textwidth]{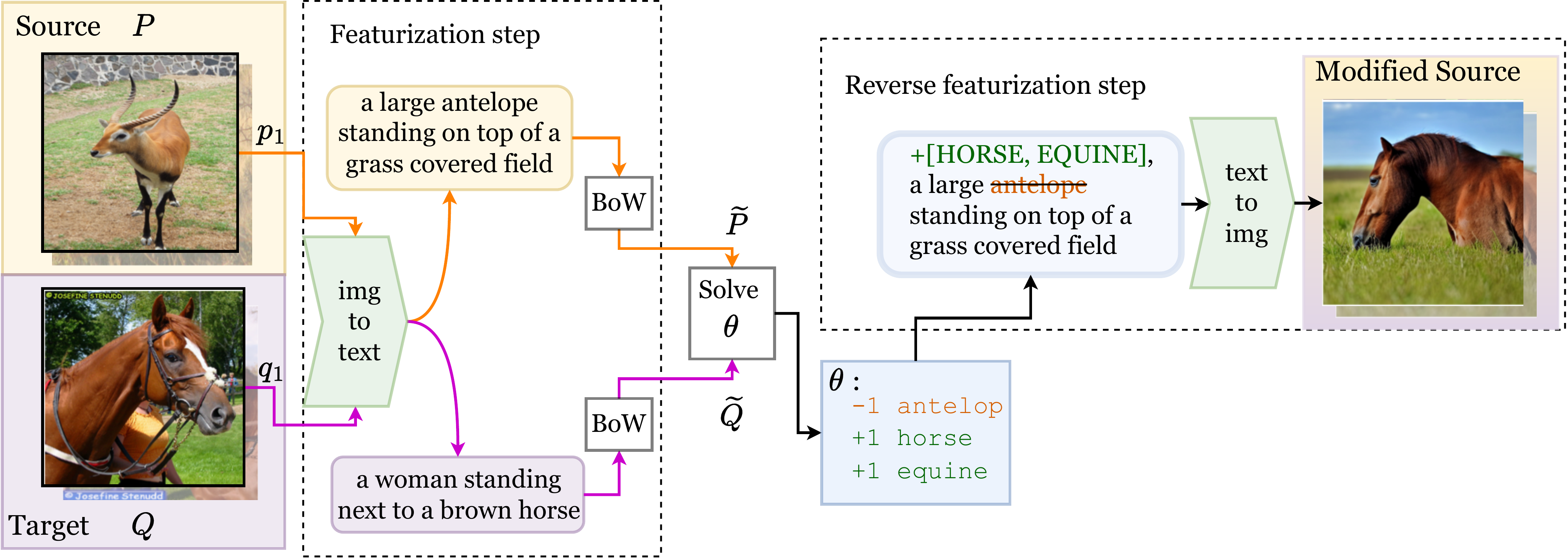}
    \caption{Pipeline of generating a shift explanation for a raw image and evaluating the explanation. 
    First, each image from the source and the target is transformed to its caption with pretrained Img2Text models (e.g., CLIP Interrogator). Then for the caption data, we derive shift explanations over interpretable features, i.e., BoW features (denoted by $\widetilde{\psrc}$ and $\widetilde{\ptgt}$ for source images and target images respectively). After deriving shift explanations over $\widetilde{\psrc}$ and $\widetilde{\ptgt}$, the modified caption is produced, which is then fed into a pretrained Text2Img model (e.g., Stable Diffusion) for reverse featurization. 
    }
    \label{fig:img_cf_overview}
    \vspace{-0.15in}
\end{figure*}

It is worth noting that shift explanations are built upon interpretable features, e.g., age or education level for the Adult dataset, which, however, are not available for image and language data. 
Therefore, we add two additional steps in our framework. The first one is a {\it featurization step}, which extracts interpretable features from the language and image data. Second, we add a {\it reverse featurization step} for converting modified features back to the raw data space for 
producing mapped source samples.

\paragraph{Generalizing to language data} For language data, the {\it featurization step} leverages techniques such as Bag-of-words (BoW) and N-Gram models to produce token-level features. These features for the source and target data are denoted by $\widetilde{\psrc}$ and $\widetilde{\ptgt}$ respectively. Then, $\widetilde{\psrc}$ and $\widetilde{\ptgt}$ can be integrated into $L(\theta; M, \psrc, \ptgt)$ and $\text{WG-}L(\theta; M, \psrc, \ptgt)$ for solving $\theta$.
The resulting mapping function $M(\cdot; \theta)$ is in the form of removal or addition of words. Therefore, in the {\it reverse featurization step}, we follow the explanations to either remove words from the sentences in the source distribution or add words to the beginning of these sentences. 

\paragraph{Generalizing to image data} In comparison to language data, both {\it featurization} and {\it reverse featurization steps} over images are even more difficult. To address this issue, we propose an end-to-end pipeline shown in Figure~\ref{fig:img_cf_overview}. 
The featurization step starts by leveraging image-to-text models such as CLIP Interrogator \citep{clipinter}
to produce captions for each image from the source distribution and the target distribution. These captions are then processed in the same manner as language data to obtain interpretable features, such as BoW features, which are denoted by $\widetilde{\psrc}$ and $\widetilde{\ptgt}$ for the source and the target respectively.
We then follow the way of generating shift explanations over language data to find shift explanations. Finally, the {\it reverse featurization step} follow the explanation to produce modified captions for each source image, which is then transformed back to an image using a text-to-image model such as stable diffusion model \citep{rombach2021highresolution}.

\subsection{Feasibility and Robustness Metrics}\label{sec: fr}

Despite varied objective functions $L(\theta; M, \psrc, \ptgt)$ across the different shift explanation methods, to our knowledge, PercentExplained (PE) from Equation \eqref{eq: pe} is the only metric to evaluate the quality of \ceds{}s in the literature \citep{kulinski2022towards}. 
We propose to use feasibility and robustness, introduced in Section~\ref{sec: problem}, as additional metrics. We formalize feasibility and robustness as evaluation metrics below.

{\bf Feasibility}
This notion of feasibility has been studied in the literature of counterfactual explanations \citep{poyiadzi2020face}. Formally speaking, feasibility is defined as the percentage of source samples for which the explanations are feasible, i.e.:
\begin{small}
\begin{align}\label{eq: feasiblity}
    \text{\% Feasible} = [{\sum\nolimits_{x\in\psrc} a(x, M(x; \theta))}]/{\|\psrc\|}
\end{align}    
\end{small}
where $a(\cdot, \cdot)$ is a function which outputs 1 when the change from $x$ to $M(x; \theta)$ is feasible, and 0 otherwise (say changing education is feasible while changing sex is almost infeasible for Adult dataset).
Evaluating feasibility becomes indispensable in the presence of unactionable attributes such as ``sex''.
Since \ourmethod{} takes groups into account, we can enhance an explanation's feasibility by constructing groups using the unactionable attributes. 

{\bf Robustness} 
The notion of robustness is also proposed in prior works such as \citep{alvarez2018robustness, agarwal2022rethinking}, which evaluates variations of the explanations with respect to small perturbation over the distribution of the source data. To add such small perturbations to the source data distribution, $\psrc$, we randomly perturb $\epsilon\%$ of the feature values for some pre-specified feature, e.g., changing the sex of 1\% of the samples from male to female. The resulting perturbed source distribution is denoted as $P(\epsilon)$.
We investigate the robustness of \ceds{}s with respect to two types of perturbations, random perturbations and worst-case perturbations. These two types of perturbations lead to two robustness metrics (denoted by $\Omega$ and $\Omega_{\text{worst}}$ respectively) which are quantified with the following formula adapted from the robustness metrics in \citep{alvarez2018robustness}):
\begin{small}
\begin{align}\label{eq: robustness_formula}
    \begin{split}
        & \Omega(\theta; M, \psrc, \ptgt, \epsilon) = {\|M(\psrc; \theta) - M(\psrc(\epsilon); \theta(\epsilon))\|_2}/{\|\psrc - \psrc(\epsilon)\|_2}, \\
        & \Omega_{\text{worst}} (\theta;M, \psrc, \ptgt) = 
        \max\nolimits_{\epsilon} \Omega (\theta;M, \psrc, \ptgt, \epsilon),
    \end{split}
\end{align}    
\end{small}

in which $\theta$ and $\theta(\epsilon)$ are derived by 
finding a shift explanation
with the source distribution as \psrc{}, or $\psrc(\epsilon)$, respectively. Since it is difficult to exactly solve $\Omega_{\text{worst}}$, we therefore follow \citep{alvarez2018robustness} to determine $\Omega_{\text{worst}}$ from a finite set of epsilons. Details are shown in Appendix \ref{appendix: fr}.

\section{Experiments}\label{sec: exp}
We present our experiments for evaluating the effectiveness of \ourmethod{} compared to shift explanations which ignore group structures. In what follows, we describe the experimental setup in Section~\ref{sec: exp_setup}, the datasets in the experiments in Section~\ref{sec: exp_dataset}, and our experimental results in Section~\ref{sec: exp_results}.

\subsection{Datasets}\label{sec: exp_dataset}%

We perform experiments on three different types of data: tabular, language, and vision data. For tabular data, we use the Adult and Breast Cancer datasets (Breast dataset for short) from the UCI Machine Learning Repository \citep{Dua:2019}. For language data, we evaluate on the \civil{} dataset \citep{borkan2019nuanced} (Civil dataset for short).
Finally, for image data we use the version of the ImageNet dataset from \citep{santurkar2021breeds}.
Appendix~\ref{appendix: datasets} provides further details of these datasets.

{\bf Distribution shift setup} For tabular data and language data, we match the setup of \citep{kulinski2022towards} and \cite{santurkar2021breeds}, and consider distribution shift between the different class labels: shift from low-income to high-income for Adult, benign to malignant for Breast, toxic to non-toxic for Civil dataset, and between sub-classes of ``Mammal'' for ImageNet. 

{\bf Sub-population setup} For the Adult dataset, we group samples by their sex attribute.
For Breast dataset, we group by the ratio between ``cell radius'' and ``cell area'' attributes (see Appendix~\ref{sec: exp_appendix} for details), leading to 3 groups.
For Civil dataset, groups are defined by samples with and without the ``female" demographic feature.
For ImageNet, groups are defined by the superclasses ``rodent/gnawer" and ``ungulate/hooved mammal" of the ImageNet label. As we show in Section \ref{sec: quantitative}, despite only a few pre-specified groups across all the datasets, the state-of-the-art shift explanations still break those group structures and lead to poor feasibility and robustness.

\subsection{Experimental Setup}\label{sec: exp_setup}
For all datasets described in Section~\ref{sec: exp_dataset}, we evaluate three shift explanation methods: $K$-cluster transport ($K$-cluster), Optimal transport (OT), and \dice{}. Due to space limitations, only the results of $K$-cluster transport are included in this section and other experiments can be found in Appendix \ref{appendix: datasets}.  For each method, we compare the vanilla explanations and those generated using \ourmethod{}. The former one are derived by optimizing group-free objectives such as PE in Equation \eqref{eq: pe} while the latter one are constructed by optimizing group-aware objectives such as WG-PE in Equation \eqref{eq: worst_group_pe}.

The three different explanation methods in addition to their counterparts using \ourmethod{} are evaluated along the following axes:
\begin{itemize}[leftmargin=*,itemsep=0pt,topsep=0pt,parsep=0pt,noitemsep]
    \item PE and WG-PE. Note that for image dataset, PE and WG-PE are evaluated over the embeddings of images rather than the raw images, produced by leveraging a pretrained ResNet-50 model.
    \item \% Feasible as shown in Equation \eqref{eq: feasiblity}. 
    \item Robustness and worst-case robustness as shown in Equation \eqref{eq: robustness_formula} by perturbing a randomly selected 1\% of the feature values for six randomly selected features. 
\end{itemize}

Recall that the \% Feasible and Robustness metrics are not differentiable, and thus used as post-hoc evaluation metrics for shift explanations.
Further details of the experimental setup are in Appendix~\ref{hyperparams}.

\subsection{Results}\label{sec: exp_results}

\begin{figure}[t]
    \centering
    \includegraphics[width=0.95\textwidth]{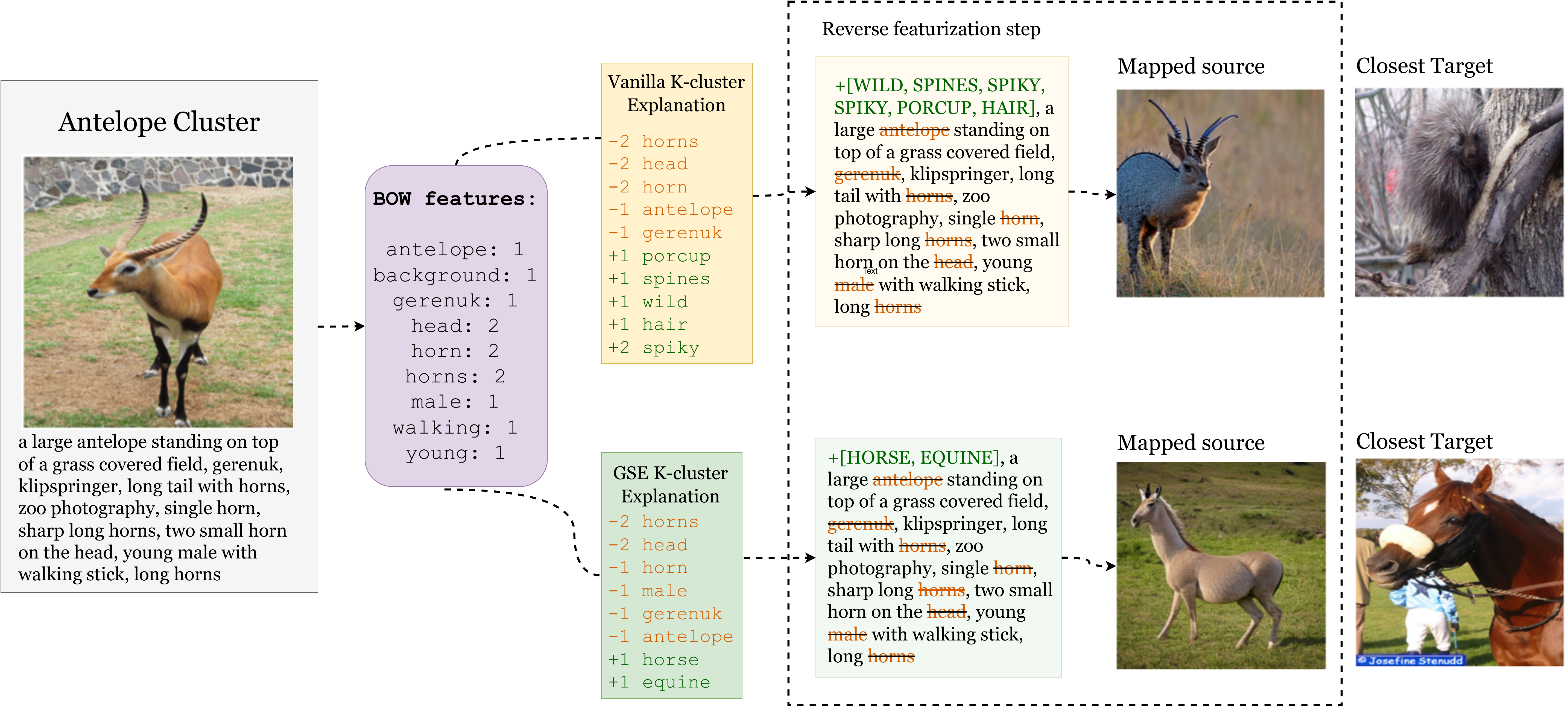}
    \caption{
    Comparison of vanilla \ceds{} and \ourmethod{} \ceds{} to explain the sub-population shift in Imagenet dataset with $k$-cluster transport. This shows how the explanation modifies samples in one cluster for the image data. The vanilla explanation maps this cluster consisting of mostly antelopes (which belong to ``ungulate/hooved mammal'' group) shown on the left to samples in the target of porcupines shown on the top right. The explanation of ``-2 horns'' and ``+2 spiky'' means that two occurrences of the word ``horns'' should be removed and ``spiky'' should be added twice to the caption. Then we apply the reverse featurization step introduced in Section \ref{fig:img_cf_overview} to this modified caption for generating mapped source image. By inspecting its closest target image, we observe that this mapped image is closer to the group ``rodent/gnawe'', thus breaking the group structure. 
    In contrast, \ourmethod{} explanation maps this cluster closer to horse images, which both belong to ``ungulate/hooved mammal'', thus preserving groups (shown on the bottom right).
    }
    \label{fig:imagenet-qualitative}
    \vspace{-0.05in}
\end{figure}

Our results are divided into quantitative and qualitative results below. We also perform experiments without prespecified groups for language data, and the results are in Appendix~\ref{no-group-exp}.
\subsubsection{Quantitative Results}\label{sec: quantitative}

\begin{table*}[t]
  \captionsetup{justification=centering} %
  \caption{Comparison of PE, WG-PE and \%Feasible between vanilla $K$-cluster explanations and \ourmethod{} $K$-cluster explanations (Higher is better). 
}
\label{main-tabular-results}
\begin{center}
\begin{small}
\begin{sc}
\begin{tabular}{lrrrrr}
\toprule
  &Dataset & Adult & Breast &  Civil & ImageNet \\
\midrule
\multirow{2}*{PE} & Vanilla &\B 24.92$\pm$1.26 & \B 85.35$\pm$0.32 & \B 12.73$\pm$0.14 & 4.46 $\pm$6.35 \\
& \ourmethod{}& 24.27$\pm$0.61&84.54$\pm$0.71& 6.23$\pm$0.54 & \B 12.25$\pm$1.96 \\
\multirow{2}*{WG-PE} & Vanilla &6.09$\pm$5.60 & 53.31$\pm$5.54& 3.85$\pm$0.02 & -16.55$\pm$5.31 \\
& \ourmethod{} &\B 21.91$\pm$2.25&\B 73.15$\pm$0.37 & \B 6.23$\pm$5.90 &\B -4.69$\pm$ 2.06  \\
\multirow{2}*{\%Feasible} & Vanilla& 84.73$\pm$10.80&\B 58.49$\pm$3.71& 57.50 $\pm$ 0.00 & 20.96 $\pm$9.36 \\
& \ourmethod{}&\B 100.0$\pm$0.0 & \B 58.49$\pm$3.71 & \B 61.83 $\pm$ 0.94 & \B 48.61$\pm$3.23 \\
\bottomrule
\end{tabular}
\end{sc}
\end{small}
\end{center}
\vskip -0.1in
\end{table*}

\begin{table*}[h]
  \captionsetup{justification=centering} 
\centering
\caption{Comparison of Robustness and Worst-case Robustness between vanilla $K$-cluster explanations and \ourmethod{} $K$-cluster explanations (Lower is better).}
\label{tabular-robustness-res}
\begin{center}
\begin{small}
\begin{sc}
\begin{tabular}{lrrrrr}
\toprule
  &Dataset & Adult & Breast &  Civil & ImageNet \\
\midrule
\multirow{2}*{Robustness} & Vanilla & 78.19 & 436.63 & 1.61& 19.65\\
& \ourmethod{}&  \B 66.52 & \B 251.86 & \B 1.39 & \B 18.20\\
\multirow{2}*{Worst-case Robustness} & Vanilla & 312.56 & 325674.59 & 17.22 & 24.18\\
& \ourmethod{} & \B 298.25 & 349549.71 & 17.36 & \B 17.84\\
\bottomrule
\end{tabular}
\end{sc}
\end{small}
\end{center}
\vskip -0.1in
\end{table*}

The main quantitative results of vanilla $K$-cluster explanations and \ourmethod{} $K$-cluster explanations are shown in Table \ref{main-tabular-results}-\ref{tabular-robustness-res}. First of all, as Table \ref{main-tabular-results} shows, for vanilla $K$-cluster explanations, a huge gap exists between the overall PE metric and WG-PE, which is up to 32\% (see Breast dataset). This thus indicates that these explanations fail to map at least one group of source samples to the target samples of the same group, thus causing group irregularity.  

By comparing \ourmethod{} explanations against the vanilla explanations, \ourmethod{} almost always results in a higher WG-PE (up to 20\% improvement, see Breast dataset) than vanilla explanations, while only slightly hurting overall PE on Adult, Breast and Civil dataset, and even improving it on ImageNet dataset. 
We also notice that \ourmethod{} always produces more feasible explanations in comparison to vanilla explanations, which improves \%Feasible by up to 28\%. This is primarily due to the fact that \ourmethod{} searches the explanations preserving groups by nature. 
Moreover, according to Table \ref{tabular-robustness-res}, \ourmethod{} improves both the robustness and worst-case robustness by up to 42\% (see the Robustness metric for Breast dataset) in almost all the cases across all the datasets. 

\subsubsection{Qualitative Results}
For a qualitative analysis of vanilla shift explanations compared to our \ourmethod{} ones, we first look at some examples of group irregularities in terms of broken and preserved groups. For image data, Figure~\ref{fig:imagenet-qualitative} shows the shift in an antelope cluster of the $K$-cluster explanation. We see that the vanilla explanation maps antelopes to porcupines which breaks the ``ungulate/hooved mammal'' group since antelopes are hooved animals while porcupines are rodents. In addition, observing the generated examples for this cluster shows that converting an antelope to a porcupine is difficult and yields unusual-looking results. On the other hand, \ourmethod{} maps this cluster of antelopes to horses which preserves the groups since horses are also hooved animals. The resulting generated images from this explanation are also clearly images of horses which explains why \ourmethod{} has higher feasibility than vanilla explanations for image data. 

We also perform qualitative analysis to understand why \ourmethod{} also improves robustness, which is included in Appendix~\ref{extra-qual}. The qualitative analysis on language and tabular data is also given in Appendix~\ref{extra-qual}.

\section{Related Work}\label{sec: related_work}
{\bf Explaining distribution shift.} \citep{kulinski2022towards} 
proposes three different mappings of varying levels of interpretability and expressiveness as shift explanations. 
A related problem concerns finding counterfactual explanations for explaining model behaviors \citep{mothilal2020explaining}.
Counterfactual explanation techniques find the minimal perturbation which changes a model's prediction on a given sample \citep{wachter2017counterfactual, changexplaining,rathi2019generating}. Although not originally created to explain distribution shift, we adapt these methods to our setting (see Appendix \ref{appendix: instantiations} for details). 
Note that none of these techniques take group structures into account. It is also worth noting that some works such as \citep{hou2021visualizing} explain how the models are adapted across distributions rather than explain the shift of a distribution itself, which is thus outside the scope of this paper.

{\bf Worst group robustness.}
Improving model robustness over sub-populations using group information is extensively studied in the robustness literature. Here, the main goal is to minimize the loss on the worst performing sub-population. This problem can be addressed by directly optimizing worst-group loss \citep{sagawa2019distributionally, zhangcoping}, re-weighting sub-populations \citep{liu2021just, byrd2019effect}, or performing data augmentation on the worst group \citep{goelmodel}. Rather than focus on improving worst-group model performance, our focus is to find explanations that preserve group structures.

{\bf Domain generalization and adaptation.}
Common solutions for dealing with distribution shift include {\it domain generalization} and {\it domain adaptation}.
We survey them in detail in Appendix \ref{appendix: related work}.
\section{Conclusion and Future Work}
We identified a problem with all existing approaches for explaining distribution shift: the blindness to group structures. Taking group structures into account, we developed a generic framework that unifies existing solutions for explaining distribution shift and allows us to enhance them with group awareness. These improved explanations for distribution shift can preserve group structures, as well as improve feasibility and robustness. We empirically demonstrated these properties through extensive experiments on tabular, language, and image settings. 

\printbibliography
\newpage
\appendix
\onecolumn

\section{Additional Framework Instantiations}\label{appendix: instantiations}
\subsection{Optimal Transport (OT)}

Similar to $K$-cluster Transport ($K$-cluster) \citep{kulinski2022towards}, Optimal Transport (OT) finds the moving distance $\theta(x)$ for shift explanations directly. In the next two sub-sections, we discuss how to instantiate $M(x;\theta)$ and $L(\theta; M, \psrc, \ptgt)$ for OT within our framework.

{\bf Mapping function for OT.} In OT, the mapping is almost the same as that for $K$-cluster except that the moving distance $\theta$ now depends on each individual sample, $x$, from the source. Therefore, the counterfactual mapping $M(x; \theta)$ can be written as $M(x_i; \theta_i) = x_i + \theta_i$ for every $x_i\in\psrc$. 

{\bf Objective function for OT.}
The objective function for OT is exactly the same as that for $K$-cluster which is the PE metric. The optimization now results in learning $\theta=\{\theta_1, \dots, \theta_{|\psrc|}\}$, or a separate moving distance for every source sample such that the PercentExplained is maximized.

\subsection{\dice}\label{sec: dice}
For vanilla counterfactual explanation methods such as \dice, model behavior for a given sample $x$ is explained. To construct such explanations, these methods perform counterfactual modifications to $x$ such that the model prediction changes. We adapt these methods to construct a surrogate shift explanation by finding counterfactual examples for models that classify between source and target distributions. In this subsection, we investigate how general methods for finding counterfactual examples can be adapted to fit within our framework. We take \dice{} as an example to describe how to instantiate $M(x; \theta)$ and $L(\theta; M, \psrc, \ptgt)$ for these methods.

{\bf Mapping function for \dice{}.}
The counterfactual examples produced by \dice{} depend on a given model (parameterized by $\theta$).
As a consequence, the mapping function for \dice{}, 
$M(x, \theta)$, is represented as $M(x, \theta) = x + f(x; \theta)$. Let $h$ denote the fixed model which classifies between the source and target data. The moving distance, $f(x;\theta)$, used in the counterfactual explanation relies on this model, $h$, that \dice{} is used to explain.

{\bf Objective function for \dice{}.} 
As indicated above, it is essential to obtain the parameter $\theta$ to learn the shift explanation.
Since the model, $h$, discriminates between the source data, \psrc{}, and the target data, \ptgt{}, we optimize the following objective function for \dice{}, in which all source samples and target samples are labeled as 0 and 1 respectively:
\begin{align}
\begin{split}
    & \argmin_{\theta} L_{\text{DiCE}}(\theta; M, \psrc, \ptgt)\\
    & = \argmin_{\theta} \sum_{x, y \in D} \ell(h(x; \theta),y).
\end{split}
\end{align}

In the above formula, the loss $\ell(\cdot)$ represents the Cross Entropy loss and $h$ denotes the model which classifies between the source and target data $D = \{(x, 0): x\in \psrc\}\cup \{(x, 1): x\in \ptgt\}$. Note that the above loss function is an instantiation of the abstract objective function, $L$, used in Equation \eqref{eq: generalized_loss}. This optimization leads to learning a $\theta$ which is the model parameter for the classifier between the source and target. Once we have learned the model parameter for the model $h$ to be explained with \dice{}, we derive the moving distance as
\begin{align}
\begin{split}
    & f(x;\theta) = \text{argmin}_{\delta_x} \text{dist}(x, x + \delta_x)\\
    & \textit{s.t.}\ h(x + \delta_x;\theta) = 1.
\end{split}
\end{align}

For any $x\in \psrc$ the moving distance $f(x; \theta)$ is found such that it is a minimal change to $x$ which results in the previously learned classifier, $h$, classifying the modified sample as a target sample.

\section{Details on Feasibility and Robustness}\label{appendix: fr}
\begin{figure}[t]
    \centering
    \begin{subfigure}[b]{0.34\textwidth}
    \centering
    \includegraphics[width=\textwidth]{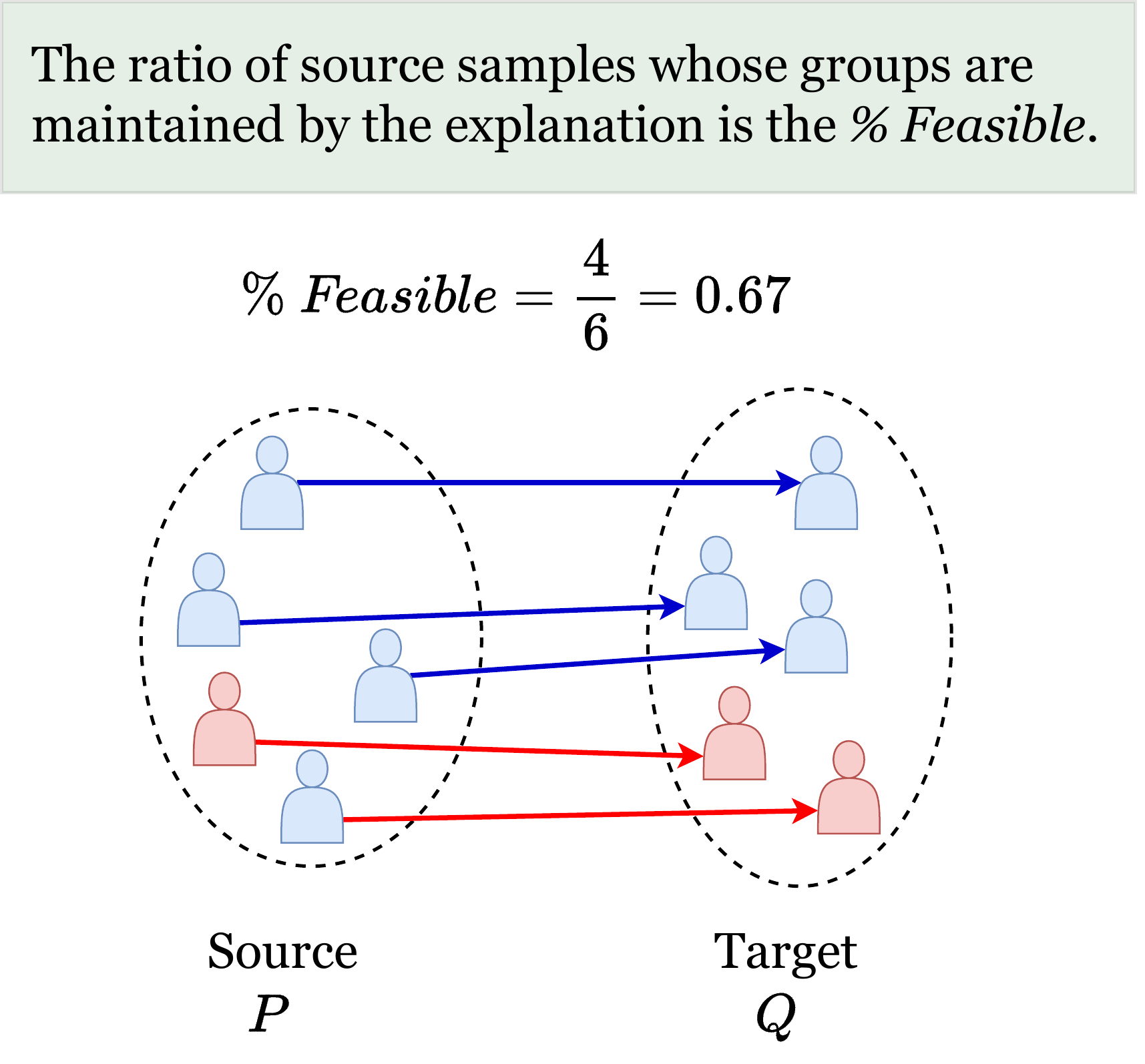}
    \caption{\% Feasible}
    \label{feas_vis}
    \end{subfigure}
    \begin{subfigure}[b]{0.64\textwidth}
    \centering
    \includegraphics[width=\textwidth]{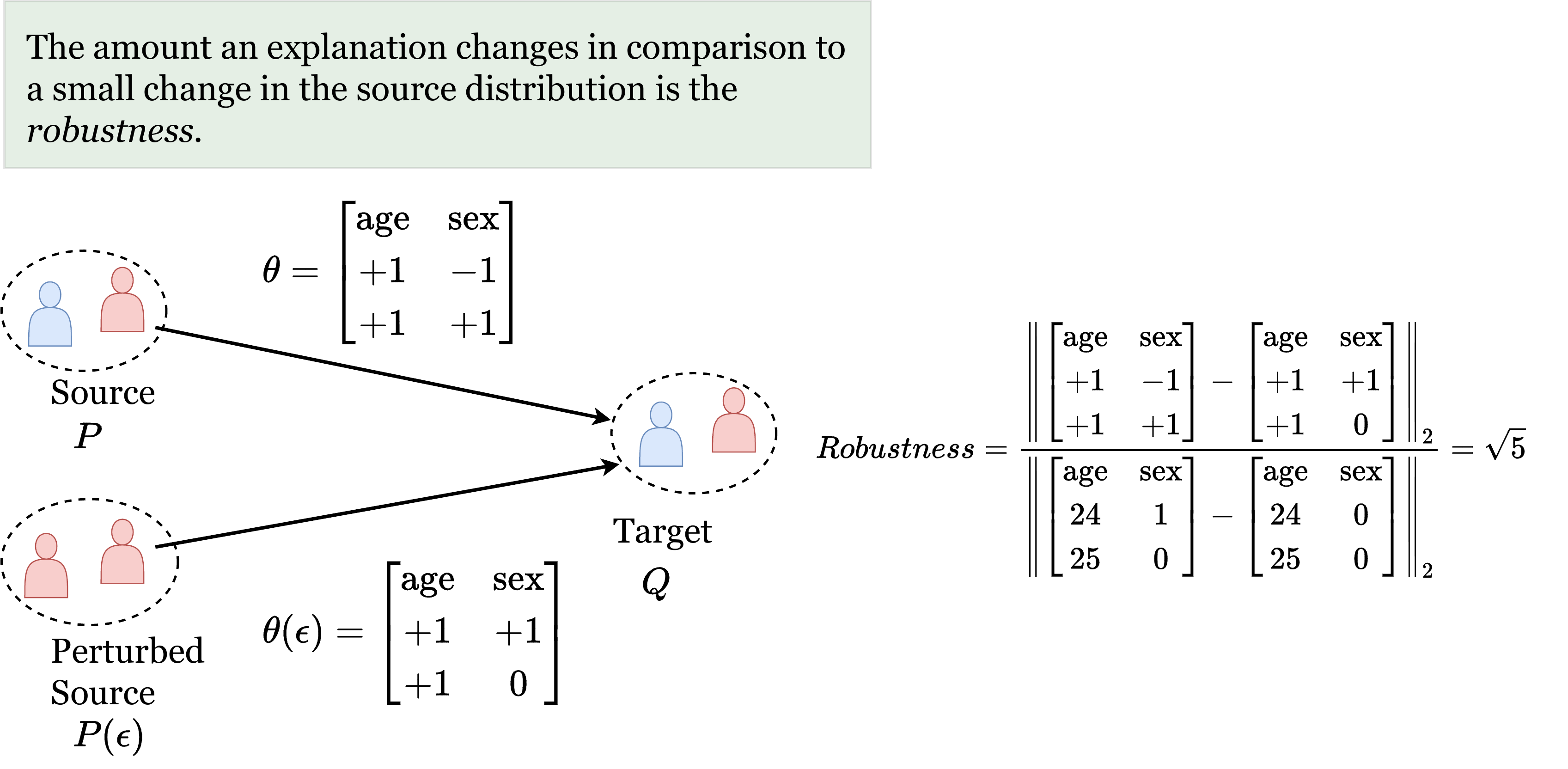}
    \caption{Robustness}
    \label{flip_vis}
    \end{subfigure}
    \caption{Visualizations of feasibility and robustness. \% Feasible is shown in Figure~\ref{feas_vis} and it measures the percent of samples which are mapped by $G(x;\theta)$ to a sample with the same group as the original sample. Robustness is shown in Figure~\ref{flip_vis} and it measures how small perturbations on the source data distribution change \ceds{}.}
    \vspace{-0.15in}
\end{figure}

Feasibility and robustness are defined in Equation~\ref{eq: feasiblity} and \ref{eq: robustness_formula} respectively, but here we give a visual example of each.
A concrete example for calculating feasibility is shown in Figure~\ref{feas_vis}. The source cluster of four males and one female becomes three males and two females from the mapping, so feasibility is $\frac{4}{6}=0.667$.

Similarly, we calculate robustness for an example in Figure \ref{flip_vis}. Suppose there are two clusters in the source distribution and the target distribution respectively, and each cluster consists of a single sample. After applying $K$-cluster transport, the moving distance $\theta$ from the source to the target can be interpreted as ``increasing the age by 1 and flipping the sex attribute''. After perturbing the sex attribute of one source sample from 1 to 0, the magnitude of changes on the source data distribution is $\|P-P(\epsilon)\|_2=1$. This produces a new moving distance $\theta(\epsilon)$, which is interpreted as ``increasing the age by 1 and only flipping the sex attribute of the first source sample''. By leveraging Equation \eqref{eq: robustness_formula}, the Robustness measure $\Omega$ for this example is $\frac{\|M(\psrc; \theta) - M(\psrc(\epsilon); \theta(\epsilon))\|_2}{\|P-P(\epsilon)\|_2} \approx \frac{\|\theta - \theta(\epsilon)\|_2}{\|P-P(\epsilon)\|_2} = \sqrt{5}$.

The details for how we produce a perturbation, calculate worst-case robustness, and perform the robustness experiment are given in Appendix~\ref{app: robustness-exp}.

\section{Datasets}\label{appendix: datasets}
The tabular, language, and image datasets that we use in the experiments are described in this section.

\subsection{Tabular data}

{\bf Dataset overview.} The Adult dataset and the Breast Cancer dataset are standard tabular datasets from the UCI Machine Learning Repository \citep{Dua:2019}. The Adult dataset consists of 48,842 samples with categorical and integer features from census data. The typical task is to predict whether income exceeds \$50K per year. The Breast Cancer dataset contains 569 samples with 10 real-valued features relating to an imaged cell. This dataset is similarly used for binary classification between the classes of benign and malignant tumors.

{\bf Distribution shift setup.} For both the Adult and Breast Cancer datasets, we match the setup by \citep{kulinski2022towards} and consider distribution shift between the different class labels: above 50k and below 50K for Adult, and benign and malignant for Breast Cancer. 

{\bf Sub-population setup.} For the Adult dataset, we use the existing demographic feature of ``male" to define two groups.
For the Breast Cancer data, we define groups by thresholding on a new attribute which is calculated by using ``cell radius'' and ``cell area'' attributes (see Appendix~\ref{sec: exp_appendix} for details).
This leads to 3 groups in total.

\subsection{Language data}
{\bf Dataset overview.} The \civil{} dataset \citep{borkan2019nuanced} is used for our language application. This dataset targets predicting the toxicity of up to 2 million public comments and it additionally contains annotations of demographic categories including gender, race, and religion of the authors of each comment. This dataset is a part of the WILDS \citep{koh2021wilds} distribution shift benchmarks and it is used to benchmark subpopulation shift. Subpopulation shift occurs when the proportions of samples from different demographic categories changes between the source and target.

{\bf Distribution shift, sub-population and featurization setup.} We build a distribution shift setting by splitting the \civil{} dataset into toxic and non-toxic text as the source and target respectively as done by \citep{kulinski2022towards}. After balancing the size of this split, there are 4,437 samples in each of the source and target. The groups are defined by samples with and without the ``female" demographic feature. The interpretable features for this data are defined by the bag-of-words representation for each sample. 
We limit the bag-of-words to 50 words which helps avoid model overfitting in \dice{} based on our observations.

\subsection{Image data}
{\bf Dataset overview.} We use BREEDS \citep{santurkar2021breeds}, which uses the wordnet class hierarchy to create subsets of ImageNet \citep{deng2009imagenet} for sub-population shift studies. We use BREEDS to create a subset of the ImageNet validation set which consists of 50,000 images. %

{\bf Distribution shift setup.} 
In BREEDS, We start at the subtree under ``mammal" in ImageNet's wordnet hierarchy and select three ImageNet classes under the superclass ``rodent/gnawer" and three classes under the superclass ``ungulate/hooved mammal" for both the source and target. These three classes for each superclass are chosen in an adversarial way according to \citep{santurkar2021breeds} to increase the level of subpopulation shift. In total, this subset consists of 298 samples in each of the source and target.

{\bf Featurization and sub-population setup.} As described in Section~\ref{lang-img-setup}, features are extracted by using an img-to-text model and then treating the caption as a bag-of-words representation. We use a total of 50 words in the bag-of-words as features.
Finally, groups are defined by the superclasses ``rodent/gnawer" and ``ungulate/hooved mammal" to encourage an explanation which does not map rodents to hooved mammals. This grouping allows us to define an infeasible explanation as one which maps rodents to hooved mammals or vice versa.

Note that for the \civil{} dataset, the groups are determined by extra annotations, which are not available for the mapped samples produced by the shift explanations. To determine the group assignments of the mapped source sample, we leverage the group annotation of the mapped sample's closest target sample as an approximated annotation.

\section{Datasets and Hyperparameters for Experiments}\label{sec: exp_appendix}
\label{hyperparams}
\subsection{Tabular data}

All categorical features in the Adult data are one-hot encoded resulting in a total of 35 features. We balance the size of both source and target distribution in which results in a total of 15,682 samples for the Adult data and 424 samples for the Breast dataset. Finally, we scale the feature values of both datasets to range from 0 to 1.

The new meta-feature that is used for grouping the Breast dataset is calculated by the expression
\[\frac{\text{Avg. cell radius}^2}{\text{Avg. cell area}},\]
and then we group the data by thresholding on this new meta-feature. To find a good threshold, we compute the meta-feature for the entire source and target dataset and get the first and third quartiles. Thus, we create three groups: samples with meta-feature value below the first quartile, between the first and third quartile, or above the third quartile.

When learning vanilla and \ourmethod{} $K$-cluster explanations for the Adult data, we use 10 clusters and optimize for 100 iterations using a learning rate of 10.0. For the Breast dataset, we use 4 clusters, 100 iterations, and a learning rate of 10.0. For OT explanations, we use a learning rate of 0.05 for Adult and 1.0 for Breast and use 100 iterations of training for both. For \dice{}, we use a neural network with a single hidden layer of size 16 as the source vs. target discriminator in the DiCE experiments. This model is trained for 100 epochs with a learning rate of 0.05 and weight decay 0.0001 for the Adult data. This network is trained for 500 epochs at learning rate 0.2 and weight decay 0.0001 for the Breast Cancer data. For \ourmethod{}, we train the neural network using group DRO \citep{sagawa*-2020-distributionallyrobust} with the same hyperparameters used in the regular training procedure.

\subsection{Language data}
For the $K$-cluster experiments, we use 4 clusters and optimize for 200 iterations using a learning rate of 20. For OT explanations, we optimize for 200 iterations using a learning rate of 0.1. Finally, for the DiCE explanations we first train a logistic regression classifier for classifying the source and target samples using 1000 epochs with learning rate of 0.5 and weight decay of 0.0001. For \ourmethod{} with \dice{}, we train this logistic regression classifier using group DRO with the same hyperparameters used in the regular training procedure.

\subsection{Image data}
For $K$-cluster explanations, we use 5 clusters and optimize for 100 iterations using a learning rate of 150.0. For OT explanations, we optimize for 100 iterations using a learning rate of 0.5. Finally, for the DiCE explanations we first train a logistic regression classifier for classifying the source and target samples using 100 epochs with learning rate of 0.1 and weight decay of 0.0001, and we use group DRO to train this classifier for \ourmethod{}.

\subsection{Framework hyperparameters}
\label{app: framework-hyp}
For all experiments, we leverage group DRO loss \citep{sagawa*-2020-distributionallyrobust} for the aggregate function $F$ in Equation~\ref{eq: generalized_loss}. We also experimented with $F(X)=\max X$ and $F(X)=\sum_{x\in X} x$ with $\lambda =0.1$. Note that for the latter $F(X)$, it is only applicable to the loss function $L$ which does not preserve the addition operation over groups, such as PE. Otherwise, Equation \eqref{eq: generalized_loss} could be rewritten as $\min_\theta \left( (1+ \lambda)\cdot L(\theta; M, \psrc{}, \ptgt{}) \right)$, which is not group-aware loss.

\subsection{Robustness experiment}
\label{app: robustness-exp}
To compute the robustness metric, we use a random small perturbation to the source distribution. To create this perturbation, we randomly select 75\% of the features and perturb 1\% of the feature values for each of these features. The manner in which we perturb this 1\% of the feature values depends on the type of the feature. For real valued features, we find the standard deviation of the feature value for the current feature we are perturbing and we randomly either add or subtract $0.05 \cdot \text{stdev}$ to 1\% of the feature values. For integer features, we randomly either add or subtract 1 to 1\% of the feature values. Finally, for boolean features, we randomly either flip the label of 1\% of the True feature values or 1\% of the False features values. For categorical features, we first convert the categories to integers such that each category is given an integer from 0 to $K$-1 where $K$ is the number of categories. This allows us to generate a perturbation for categorical features in the same way as for integer features.

We use the same hyperparameters as above for learning each shift explanation on the perturbed distribution. To speed up the experiments, we first train the shift explanation on the original source distribution and then initialize the parameters of the shift distribution with the parameters learned from the original source distribution when learning the shift explanation for the perturbed distribution.

For computing the robustness metric, we use three random perturbations as described above and average the robustness over the three runs. To compute worst-case robustness, we calculate robustness from 100 random perturbations and take the worst (highest) value of robustness from the 100 trials. Since each calculation of robustness requires learning a shift explanation using the vanilla method and \ourmethod{}, this experiment is time consuming, so we don't report error bars for the worst-case robustness.

\subsection{Compute details}
For all experiments, we use a local server with four Nvidia 2080 Ti GPUs and 80 Intel Xeon Gold 6248 CPUs. Each experiment required around 2 GB of GPU memory.

\section{Additional Qualitative Analysis}
\label{extra-qual}

\section{Experiments Without Group Labels}
\label{no-group-exp}

It is possible that group labels are not always available for a dataset, but we can still use either pretrained models to extract attributes to use for defining groups or use unsupervised methods for grouping the data. We perform an experiment on the language data to show that our group-aware method is still applicable even without group supervision.

To get groups for the language data, we cluster the sentence embeddings of our source and target data. The sentence embeddings are from a state-of-the-art sentence embedding model, \verb|all-mpnet-v2|\footnote{\url{https://huggingface.co/sentence-transformers/all-mpnet-base-v2}}, and we use K-means clustering with 10 clusters to get 10 groups for the source and target. Experimental results are shown in Table~\ref{no_group_table}, and we see the same trends as for the experiments with specified groups. In particular, our group-aware explanation always results in higher worst-group PE and \% Feasible than the regular explanation. The most significant improvement in WG-PE is seen for the OT explanation with a change from 63.07\% to 93.48\%. Interestingly, we also see that our group-aware explanation has slightly improved overall PE over the vanilla \dice{} and $K$-cluster explanations.

\begin{table}[t]
\caption{Comparison of distribution shift explanation methods on language data without groups given.}
\label{nlp-results}
\vskip 0.15in
\begin{center}
\small
\begin{tabular}{lrrrrr}
\toprule
Method & PE & WG-PE  & \% Feas. & Robustness & Worst-case Robustness\\
\midrule
\dice    & $1.08\pm 0.1$ & $-5.84\pm 0.5$ & 54.50 $\pm$  0.82 & 7.82 $\pm$ 0.01  & 7.93\\
\ourmethod{} \dice & \B 14.32 $\pm$ 0.9 & \B 7.52 $\pm$ 0.5 & \B 56.33 $\pm$ 0.85 & \B 7.71 $\pm$ 0.08  & \B 7.92\\
\midrule
$K$-cluster    & 5.19 $\pm$ 1.75 & 2.64 $\pm$ 0.34 & 66.00 $\pm$ 0.71 & \B 3.00 $\pm$ 0.20 & 4.60\\
\ourmethod{} $K$-cluster & \B 5.72 $\pm$ 0.88 & \B 3.79 $\pm$ 0.27 & \B 67.00 $\pm$ 0.41 & 3.02 $\pm$ 0.05 & \B 3.21\\
\midrule
OT    & \B 99.89 $\pm$ 0.00 & 63.07 $\pm$ 2.97 & 55.17 $\pm$ 4.11 & \B 1.00 $\pm$ 0.02 & \B 1.05\\
\ourmethod{} OT & 98.34 $\pm$ 0.28 & \B 93.48 $\pm$ 0.26 & \B 84.67 $\pm$ 0.24 & 1.06 $\pm$ 0.03 & 1.14\\
\bottomrule
\end{tabular}
\end{center}
\label{no_group_table}
\vskip -0.1in
\end{table}

\section{Results for OT and \dice{} Shift Explanations}\label{appendix: all_results}
The full results for tabular data, \civil{}, and ImageNet are given in Table~\ref{tabular-results}, \ref{nlp-results}, and \ref{image-results} respectively. With \dice{} and OT shift explanations, we see the same trends as previously mentioned in relation to $K$-cluster explanations. In particular, WG-PE is always improved by \ourmethod{}, and feasibility and robustness are improved in most cases.

\begin{table*}[h]
\caption{Comparison of distribution shift explanation methods on tabular datasets. 
}
\label{tabular-results}
\vskip 0.15in
\centering
\small
\begin{subtable}[c]{\textwidth}
\centering
\begin{tabular}{lrrrrr}
\toprule
 Method & PE & WG-PE & \% Feas & Robustness & Worst-case Robustness\\
\midrule

Vanilla \dice{} & 2.25 $\pm$ 0.29 & 2.25 $\pm$ 0.24 & 100.0 $\pm$ 0.00 & 23.74 $\pm$ 4.05 & 41.58\\
\ourmethod{} \dice{} & \B 26.02 $\pm$ 3.00 & \B 21.69 $\pm$ 4.77 & 100.0 $\pm$ 1.52 & \B 22.34 $\pm$ 1.29 & \B 34.56\\
\midrule
Vanilla OT & \B 95.56 $\pm$ 0.25 & 61.23 $\pm$ 1.06 & 81.00 $\pm$ 0.36 & 54.86 $\pm$ 3.46 & 71.65\\
\ourmethod{} OT & 76.55 $\pm$ 0.19 & \B 76.55 $\pm$ 0.19 & \B 100.0 $\pm$ 0.00 & \B 47.01 $\pm$ 3.40 & \B 65.95\\
\bottomrule
\end{tabular}
\subcaption{Adult data}
\end{subtable}

\begin{subtable}[c]{\textwidth}
\centering
\begin{tabular}{lrrrrr}
\toprule
 Method & PE & WG-PE & \% Feas & Robustness & Worst-case Robustness\\
\midrule

Vanilla \dice{} & 29.6 $\pm$ 2.43 & 25.16 $\pm$ 1.00 & 25.94 $\pm$ 0.00 & 201.00 $\pm$ 21.56 & 566483.01\\
\ourmethod{} \dice{} &\B 38.21 $\pm$ 1.58 & \B 33.48 $\pm$ 0.40 & \B 27.20 $\pm$ 1.46 & \B 189.93 $\pm$ 28.37 & \B 566201.54\\
\midrule
Vanilla OT & \B 93.10 $\pm$ 0.06 & 85.79 $\pm$ 0.12 & \B 66.98 $\pm$ 0.66 & 190.02 $\pm$ 191.10 & \B 762966.00\\
\ourmethod{} OT & 90.54 $\pm$ 0.02 & \B 89.10 $\pm$ 0.07 & 61.48 $\pm$ 1.24 & \B 149.13 $\pm$ 160.04 & 819351.59\\
\bottomrule
\end{tabular}
\subcaption{Breast data}
\end{subtable}
\vskip -0.1in
\end{table*}

\begin{table}[h]
\caption{Comparison of distribution shift explanation methods on language data.}
\label{nlp-results}
\vskip 0.15in
\begin{center}
\small
\begin{tabular}{lrrrrr}
\toprule
Method & PE & WG-PE  & \% Feas. & Robustness & Worst-case Robustness\\
\midrule
\dice & 2.75 $\pm$ 0.19 & 1.11 $\pm$ 0.30 &  63.33 $\pm$ 1.25 & 5.28 $\pm$ 1.72 & 6.75\\
\ourmethod{} \dice & \B 19.29 $\pm$ 0.80 & \B 15.12 $\pm$ 2.47 &  \B 64.67 $\pm$ 0.62 & \B 1.72 $\pm$ 0.06 & \B 3.40\\
\midrule
OT & \B 99.89 $\pm$ 0.09 & 74.67 $\pm$ 1.19 &  49.83 $\pm$ 3.86 & 4.30 $\pm$ 0.24 & 5.81\\
\ourmethod{} OT & 94.62 $\pm$ 1.12 & \B 94.62 $\pm$ 1.12 &  \B 60.67 $\pm$ 0.24 & \B 4.05 $\pm$ 0.30 & \B 5.40\\
\bottomrule
\end{tabular}
\end{center}
\vskip -0.1in
\end{table}

\begin{table}[h]
\caption{Comparison of domain shift explanation methods on image data.}
\label{image-results}
\vskip 0.15in
\begin{center}
\small
\begin{tabular}{lrrrrr}
\toprule
Method & PE & WG-PE  & \% Feas. & Robustness & Worst-case Robustness\\
\midrule
\dice & -1.09 $\pm$ 1.54 & -17.25 $\pm$ 2.55 & \B 50.39 $\pm$ 0.42 & \B 5.08 $\pm$ 0.36 & 16.24\\
\ourmethod{} \dice & \B 0.19 $\pm$ 1.63 & \B -15.27 $\pm$ 3.08 & 49.94 $\pm$ 0.32 & 6.39 $\pm$ 0.64 & \B 15.73\\
\midrule
OT & 7.18 $\pm$ 1.04 & -17.30 $\pm$ 2.74 & 36.12 $\pm$ 0.55 & 18.77 $\pm$ 2.56 & 24.33\\
\ourmethod{} OT & \B 12.81 $\pm$ 1.34 & \B -14.70 $\pm$ 2.50 & \B 48.16 $\pm$ 0.72 & \B 7.76 $\pm$ 1.73 & \B 22.79\\
\bottomrule
\end{tabular}
\end{center}
\vskip -0.1in
\end{table}

\section{Additional Related Work}\label{appendix: related work}
{\bf Domain generalization and adaptation.}
Common solutions for dealing with distribution shift include {\it domain generalization} and {\it domain adaptation}. Domain generalization assumes that the target distribution is unknown and the goal is to improve model robustness to unseen out-of-distribution data. In contrast, domain adaptation aims to adapt a model learned on the source distributions to some {\it known target distribution}. But similar techniques were proposed for domain generalization and domain adaptation, including augmenting training data \citep{li2021simple, yao2022improving, motiian2017few}, adding regularization terms to the loss function \citep{zhao2020domain, balaji2018metareg, kim2021selfreg, cicek2019unsupervised, saitoadversarial}
and meta-learning \citep{li2018learning, motiian2017few}. There are also many real world distribution shift datasets such as the iWildCam dataset \citep{beery2021iwildcam} and the Camelyon17 dataset of \citep{bandi2018detection} as part of the WILDS datasets \citep{koh2021wilds}.

\section{Limitations and Societal Impacts}

\ourmethod{} explanations are only as good as the underlying shift explanation method. For instance, $K$-cluster transport can result in weak explanations that minimally reduce the Wasserstein distance between the source and target distributions if too few clusters are used (i.e. $K$ is chosen too small). On the other hand, the Optimal Transport explanation that we found reduced the Wasserstein distance the most, is not very interpretable since each source sample can be mapped differently. This results in the explanation being interpretable only on a per-sample basis. Improved interpretability of shift explanations is an area for future work.

In addition to interpretability of the explanation, our shift explanations for image and language data rely on interpretable feature extraction methods and methods for counterfactual modification based on changes to the features as described in Section~\ref{lang-img-setup}. We designed a system for interpretable feature extraction which uses a bag-of-words feature representation, but this method looses the context that words are used in and it is difficult to make counterfactual modifications. Creating disentangled embedding spaces for interpretable embeddings that can also be used for counterfactual modification is an area of active research, but there is still work left to make these approaches more general.

We also found that \ourmethod{} is sensitive to the choice of groups. Even though unsupervised methods can be used to select groups as shown in Appendix~\ref{no-group-exp}, future work can look at how to best select or design groups. For instance, it may be the case that we know of some groups, but we want the rest of the data to be grouped appropriately.

Finally, while we evaluated the worst-case robustness, our method sometimes results in worse worst-case robustness than the vanilla approach. This is again due to the choice of groups. Future work should investigate how to extend group robustness to worst-case group robustness of shift explanations so that a bad choice of groups does not negatively impact robustness.

Explanations which look plausible but are actually wrong can be harmful. This creates the illusion of understanding, and this can have serious downstream implications especially if policies are constructed from a shift explanation. With this work we hope to uncover some properties that a good shift explanation should have and design metrics and learning procedures based on group robustness to rectify these issues. 

\end{document}